\documentclass{article}

\usepackage{microtype}
\usepackage{graphicx}
\usepackage{subcaption}
\usepackage{booktabs} % for professional tables

\usepackage{hyperref}

\usepackage[preprint]{icml2026}

\usepackage{amsmath}
\usepackage{amssymb}
\usepackage{mathtools}
\usepackage{amsthm}

\usepackage[capitalize,noabbrev]{cleveref}

\theoremstyle{plain}

\theoremstyle{definition}

\theoremstyle{remark}

\usepackage{algorithm}
\usepackage{algorithmic}
\usepackage[utf8]{inputenc} % allow utf-8 input
\usepackage[T1]{fontenc}    % use 8-bit T1 fonts
\usepackage{url}            % simple URL typesetting
\usepackage{booktabs}       % professional-quality tables
\usepackage{amsfonts}       % blackboard math symbols
\usepackage{nicefrac}       % compact symbols for 1/2, etc.
\usepackage{microtype}      % microtypography
 \usepackage{graphicx}
\usepackage{amsmath}
\usepackage{amssymb}
\usepackage{mathtools}
\usepackage{amsthm}
\usepackage{multirow}
\usepackage{multicol}
\usepackage{algorithm}
\usepackage{algorithmic}
\usepackage{graphicx}
\usepackage{array}
\usepackage{xcolor,colortbl}
\usepackage{xcolor}         % colors

\usepackage{enumitem} % 
\usepackage{wrapfig}
\usepackage{multirow}
\usepackage{arydshln}
\usepackage{bm}
\usepackage{bbm}
\usepackage{multirow}
\usepackage{mathtools}
\usepackage{amssymb} %\blacktriangleright
\usepackage{pifont}
\definecolor{tabhighlight}{HTML}{e5e5e5}
\usepackage[capitalize,noabbrev]{cleveref}

\definecolor{ForestGreen}{rgb}{0, 0.69, 0.31}
\definecolor{NavyBlue}{rgb}{0, 0.44, 0.75}
\newcommand{\hgreen}[1]{\textcolor{ForestGreen}{\textbf{#1}}} % highlight color
\usepackage{pifont}

\definecolor{Cerulean}{rgb}{0.0, 0.48, 0.65}

\usepackage{newfloat}
\usepackage{listings}
\DeclareCaptionStyle{ruled}{labelfont=normalfont,labelsep=colon,strut=off} % DO NOT CHANGE THIS
\floatstyle{ruled}
\newfloat{listing}{tb}{lst}{}
\floatname{listing}{Listing}

\usepackage[textsize=tiny]{todonotes}

\begin{document}

\twocolumn[
  \icmltitle{Learning Domain Knowledge in Multimodal Large Language Models through Reinforcement Fine-Tuning}



  

  \begin{icmlauthorlist}
    \icmlauthor{Qinglong Cao}{1,2}
    \icmlauthor{Yuntian Chen}{2}
    \icmlauthor{Chao Ma}{1}
    \icmlauthor{Xiaokang Yang}{1}
  \end{icmlauthorlist}

  \icmlaffiliation{1}{MoE Key Lab of Artificial Intelligence, AI Institute, Shanghai Jiao Tong University, Shanghai, China}
  \icmlaffiliation{2}{Eastern Institute of Technology, Ningbo, China}

  \icmlcorrespondingauthor{Yuntian Chen}{ychen@eitech.edu.cn}

  \icmlkeywords{Machine Learning, ICML}

  \vskip 0.3in
]



\printAffiliationsAndNotice{}  % no special notice (required even if empty)

\begin{abstract}
% Multimodal large language models (MLLMs) have demonstrated exceptional performance in
% multimodal perception and understanding tasks. However, applying the MLLMs to specialized domains like remote sensing and medical domains remains relatively unexplored. Injecting domain knowledge is a critical and plausible approach for domain fine-tuning. However, we observe that naively injecting domain knowledge through textual conditioning or captioning does not yield obvious improvements on scientific multimodal tasks. This suggests that current MLLMs still lack domain-internalized priors and that such knowledge must be integrated at the optimization level rather than through input prompts or instructions. To address this, we propose a domain knowledge-based reinforcement fine-tuning framework that incorporates domain knowledge directly into the learning objective. Concretely, we are adding the domain-knowledge inspired constraints, and adjusting the advantages based on the domain knowledge measurement, also rewarding the output answer in the domain space.  We conduct extensive experiments on several domain-specific datasets, demonstrating the state-of-the-art performance and highlighting the effectiveness of optimization-level domain knowledge injection.

Multimodal large language models (MLLMs) have shown remarkable capabilities in multimodal perception and understanding tasks. However, their effectiveness in specialized domains, such as remote sensing and medical imaging, remains limited. A natural approach to domain adaptation is to inject domain knowledge through textual instructions, prompts, or auxiliary captions. Surprisingly, we find that such input-level domain knowledge injection yields little to no improvement on scientific multimodal tasks, even when the domain knowledge is explicitly provided. This observation suggests that current MLLMs fail to internalize domain-specific priors through language alone, and that domain knowledge must be integrated at the optimization level. Motivated by this insight, we propose a reinforcement fine-tuning framework that incorporates domain knowledge directly into the learning objective. Instead of treating domain knowledge as descriptive information, we encode it as domain-informed constraints and reward signals, shaping the model’s behavior in the output space.
Extensive experiments across multiple datasets in remote sensing and medical domains consistently demonstrate good performance gains, achieving state-of-the-art results on multimodal domain tasks. Our results highlight the necessity of optimization-level domain knowledge integration and reveal a fundamental limitation of textual domain conditioning in current MLLMs.

\end{abstract}

\section{Introduction}

\begin{figure}[t!]
	\begin{center}
		%\fbox{\rule{0pt}{2in} \rule{0.9\linewidth}{0pt}}img
    \includegraphics[width=0.50\textwidth]{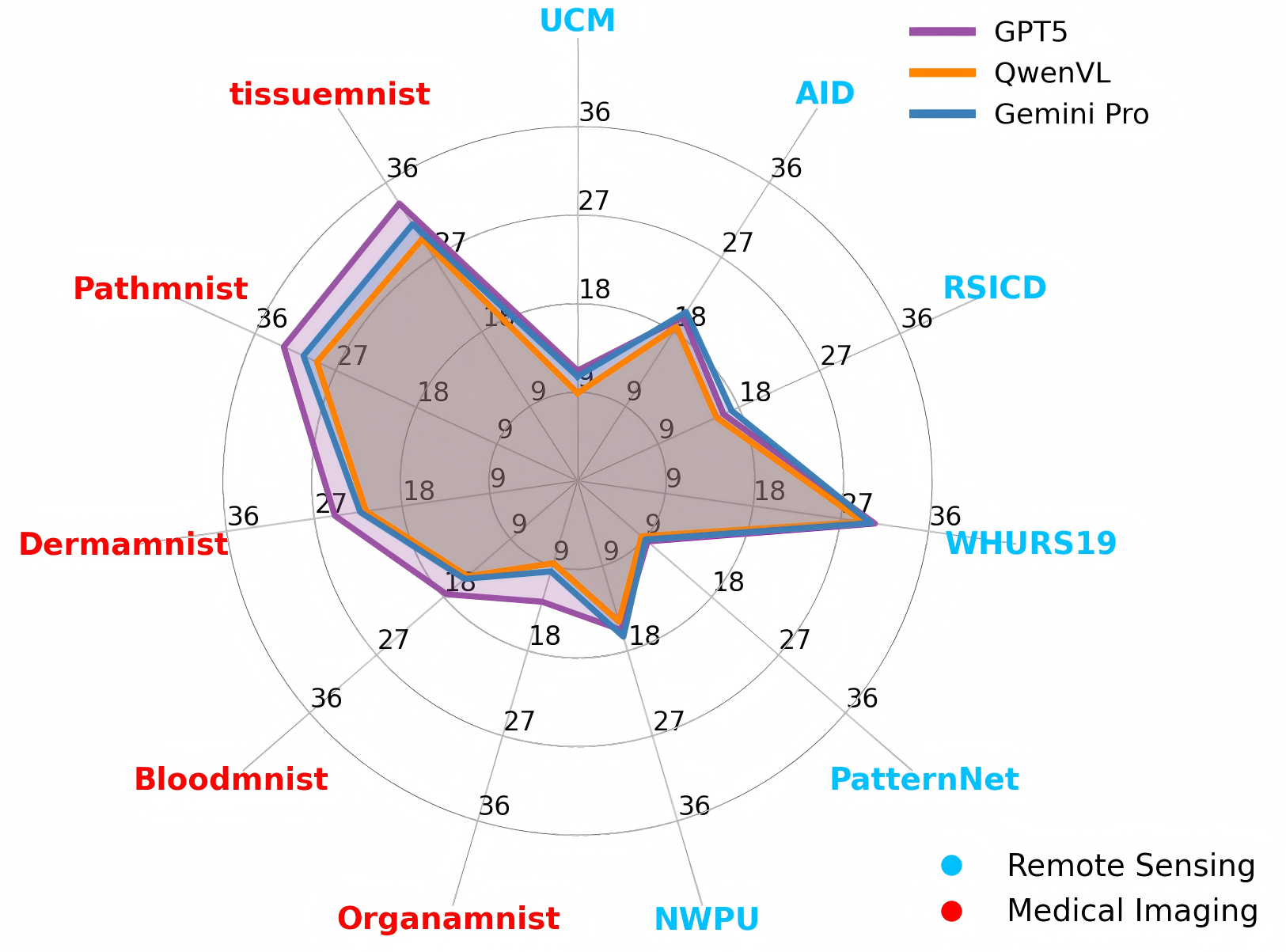}
	\end{center}
    \vspace{-2mm}
	\caption{Performance comparison in specialized domains with multimodal large language models.}
	\label{fig1}
    \vspace{-4mm}
\end{figure}

Multimodal large language models (MLLMs) have demonstrated strong generalization abilities across a wide range of vision--language tasks~\cite{achiam2023gpt,bai2025qwen2,lu2025ovis2,xu2025probing,cao2025generalized}, leading to remarkable progress in diverse application domains~\cite{jiang2025corvid,wang2025triplets,sunparrot,jiang2025multimodal}. 
Despite these successes, their effectiveness in specialized scientific domains remains limited. 
As illustrated in Figure~\ref{fig1}, advanced MLLMs typically achieve only 20\%--30\% accuracy on benchmarks from remote sensing and medical imaging, indicating substantial challenges in cross-domain generalization.

\begin{table}[t!]
    % \scriptsize
    % \footnotesize
    \centering
    \renewcommand{\arraystretch}{1.5}
    \renewcommand{\tabcolsep}{1.4mm}
    \scalebox{0.75}{
    \begin{tabular}{lcccccccc}
    \toprule
    & UCM & AID & RSICD & WHURS19 & PatternNet & NWPU   \\ 
    \midrule
    % \multicolumn{9}{c|}{\textbf{\textit{RBC fluid simulation}}} \\ \midrule
    QwenVL & 8.91	&18.6	&15.5	&29.3	&8.62	&14.9  \\ 
    + Domain Prompt & 8.89	&18.5	&15.4	&29.1	&8.71 &14.9 \\ 
    + Caption(MLLM) & 6.42  &15.8   &15.3   &26.1   &5.23 &10.5 \\ 
    + Caption(BLIP) & 8.62 & 18.1 &15.2 &28.8 &8.52 &14.7   \\ 
        \rowcolor{tabhighlight}
    Ours & \textbf{28.8}	&\textbf{38.9}	&\textbf{26.4}	&\textbf{44.4}	&\textbf{19.1}	&\textbf{36.4}
    \\
    \bottomrule
    \end{tabular}}
    \caption{Performance comparison by injecting domain knowledge with different fine-tuning methods.}
    \label{tab1}
    \vspace{-6mm}
\end{table}

A natural strategy to address this limitation is to explicitly inject domain knowledge into MLLMs through prompts or auxiliary textual descriptions. Thus, we systematically investigate such approaches in the remote sensing domain. 
Specifically, we explore (i) domain-aware prompting that explicitly encodes domain properties (e.g., rotation invariance), and (ii) caption-based augmentation, where auxiliary captions are either generated by MLLMs themselves or provided by an external captioning model (BLIP~\cite{li2023blip}). 
However, as shown in Table~\ref{tab1}, these naive forms of domain knowledge injection do not consistently improve performance. 
In several cases, caption-based augmentation even degrades accuracy, suggesting that simply providing additional domain-related text is insufficient and may introduce misleading or noisy signals.

These observations suggest that current MLLMs are unable to directly internalize high-level domain knowledge through naive textual conditioning. 
A key challenge is that domain knowledge is often abstract and conceptual rather than instance-level, and therefore cannot be easily translated into large-scale supervised annotations. 
Moreover, for specialized domains, such annotations are typically expensive and difficult to acquire. 
As a result, conventional supervised fine-tuning (SFT)~\cite{dong2024abilities,zhang2025domain,yuan2024self}, which relies heavily on explicit input–output pairs, is insufficient for effectively injecting domain knowledge into the model.

In contrast, reinforcement learning~\cite{guo2025deepseek,zhang2024large,wang2024rl,yue2025does}, which does not require dense supervision, provides a natural mechanism for incorporating abstract domain principles at the optimization level. 
By defining reward signals that reflect domain-specific constraints or desiderata, reinforcement learning enables models to gradually align their reasoning behaviors with domain knowledge, even in the absence of large-scale annotations.

% In contrast, domain knowledge in many specialized domains often manifests as abstract principles, such as invariance, consistency, or preference relations, rather than explicit paired supervision. Such knowledge is inherently difficult to encode through additional annotations or input-level prompts, as it does not directly specify correct predictions but instead constrains the desired behavior of the model across transformed or related inputs.
% Reinforcement learning~\cite{guo2025deepseek,zhang2024large,wang2024rl,yue2025does} naturally aligns with this property by operating at the optimization level, where domain principles can be expressed as reward signals or constraints.
% By shaping the policy through these signals, reinforcement learning enables models to gradually internalize domain knowledge in their reasoning behaviors, even in the absence of dense annotations.

Motivated by this insight, we propose a reinforcement fine-tuning framework that integrates domain knowledge directly into the learning objective, as show in Table~\ref{tab1}, enabling MLLMs to acquire domain-aware reasoning capabilities beyond what is achievable with prompt-based or supervised approaches. 
Specifically, we introduce a domain-aware constraint loss that explicitly encodes domain knowledge as optimization constraints. 
This constraint loss is incorporated into the reinforcement learning process, shaping the model’s policy toward domain-consistent reasoning behaviors. 
Rather than relying on explicit supervision, the proposed constraint guides the model to satisfy abstract domain principles during optimization.

In addition, we quantify the degree of domain relevance for each training sample and leverage this information to reweight the reinforcement learning advantages. 
By assigning higher advantages to samples that exhibit stronger domain-specific characteristics, the learning process emphasizes domain-aware trajectories and suppresses spurious or domain-agnostic behaviors. 
Together, the constraint loss and the advantage reweighting mechanism jointly constitute a domain knowledge–aware reinforcement signal, enabling effective domain knowledge injection even under limited annotations.

Through this unified framework, domain knowledge is no longer treated as external conditioning, but is internalized as part of the model’s optimization objective, resulting in robust and generalizable domain-aware reasoning. We conduct extensive evaluations across diverse datasets in both remote sensing and medical imaging domains. The results consistently validate the effectiveness of the proposed framework, demonstrating its ability to induce domain-aware reasoning behaviors across heterogeneous multimodal tasks.
In summary, we make the following contributions:
\vspace{-2mm}
\begin{itemize}

\item To the best of our knowledge, we first proposed a reinforcement fine-tuning paradigm that explicitly injects domain knowledge into multimodal large language models at the optimization level, moving beyond prompt-based or supervised domain adaptation.

\item We develop a domain-aware reinforcement learning framework that integrates domain knowledge directly into the learning objective through domain-specific constraints and reward shaping, together with a domain knowledge–aware advantage reweighting strategy that emphasizes domain-relevant samples during training.

\item We demonstrate the effectiveness of the proposed framework through extensive experiments on remote sensing and medical imaging benchmarks, achieving state-of-the-art performance across a wide range of multimodal domain-specific tasks.

\end{itemize}

\section{Related Work}

\noindent \textbf{Multimodal Large Language Models.} 
Multimodal large language models (MLLMs), such as GPT-4V~\cite{achiam2023gpt} and Qwen-VL~\cite{bai2025qwen2}, have demonstrated remarkable capabilities in jointly understanding and reasoning over visual and textual inputs. 
Recent advances show that MLLMs can effectively serve as general-purpose multimodal agents, enabling a wide range of downstream tasks that require vision--language perception, interaction, and decision making~\cite{li2024seed,huang2024large,han2024multimodal}. 
These models have been successfully applied to diverse multimodal applications, including visual editing, multimodal reasoning, and interactive response generation~\cite{huang2024smartedit,zang2025contextual,luo2025mmevol,wang2025divide}. From a training perspective, the development of MLLMs typically consists of two stages. 
The first stage is large-scale multimodal pre-training~\cite{luo2025mono,lin2024vila,fan2024pre}, which leverages massive image--text data to endow the model with general multimodal representations and broad world knowledge. 
The second stage is post-training~\cite{cheng2024domain,wang2024q,cheng2025domain}, which includes instruction tuning, reinforcement learning, and alignment techniques, aiming to improve instruction following, response quality, and reasoning behavior. 
While these post-training strategies have proven effective for general-purpose tasks, their ability to inject domain knowledge into MLLMs remains limited, especially in specialized scientific domains.

\noindent \textbf{Reinforcement Learning for MLLMs.} 
With the emergence of reasoning-oriented models such as OpenAI-o1~\cite{jaech2024openai} and DeepSeek-R1~\cite{liu2024deepseek,guo2025deepseek}, increasing attention has been devoted to leveraging reinforcement learning (RL) to enhance the reasoning capabilities of MLLMs. 
Reinforcement learning provides a flexible optimization framework that allows models to be trained with preference signals or task-level objectives, without relying on dense supervised annotations.
Conventional RL-based alignment methods include Proximal Policy Optimization (PPO)~\cite{schulman2017proximal}, which optimizes policies under a KL-regularized constraint, and Direct Preference Optimization (DPO)~\cite{rafailov2023direct} which eliminates the need for an explicit reward model by directly learning from pairwise preferences. 
More recently, Group Relative Policy Optimization (GRPO)~\cite{shao2024deepseekmath} has been proposed as an efficient alternative that estimates advantages using group-wise comparisons, significantly reducing computational overhead while maintaining stable training dynamics. 
Due to its favorable efficiency and scalability, GRPO has become a widely adopted reinforcement learning paradigm for training large language models~\cite{zhang2025r1,zhang2025critique,ramesh2024group,gao2024rebel}. Particularly, Visual-RFT~\cite{Liu_2025_ICCV} adopts it for visual perception tasks.
In this work, we adopt GRPO as the underlying reinforcement learning framework. 
By integrating domain-aware constraints and domain knowledge–guided reward shaping into the GRPO objective, our approach enables reinforcement learning to explicitly guide MLLMs toward domain-consistent reasoning behaviors.

\section{Method}
\begin{figure*}[h]
	\begin{center}
		%\fbox{\rule{0pt}{2in} \rule{0.9\linewidth}{0pt}}img
    \includegraphics[width=0.9\textwidth]{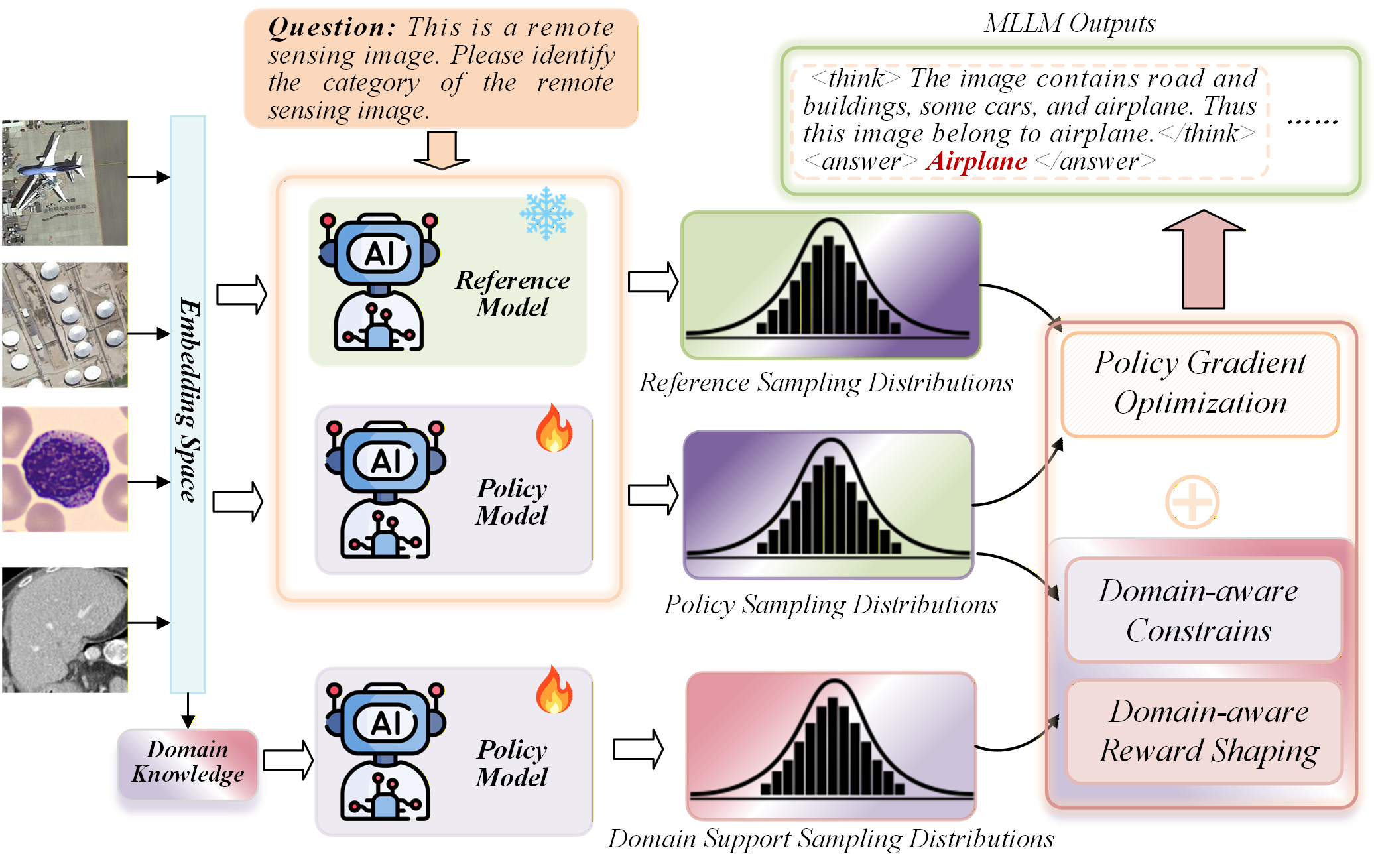}
	\end{center}
        \vspace{-3mm}
    \caption{Overview of the proposed domain-aware reinforcement fine-tuning framework. Unlike prompt-based or caption-based domain adaptation, domain knowledge is incorporated at the optimization level through domain-aware constraints and reward shaping. By modifying the policy sampling distributions under reinforcement learning, the model is guided toward domain-consistent reasoning behaviors without relying on dense annotations.}
	\label{fig2}
    \vspace{-3mm}
\end{figure*}

% \subsection{Method overview}
As illustrated in Figure~\ref{fig2}, our goal is to adapt multimodal large language models (MLLMs) to specialized domains such as remote sensing and medical imaging.
Unlike conventional supervised fine-tuning (SFT) or prompt-based domain injection, our approach internalizes domain knowledge through reinforcement fine-tuning.
We focus on abstract and structural domain knowledge that cannot be easily expressed via textual supervision, such as rotation invariance in remote sensing and symmetry consistency in medical imaging.
To incorporate such knowledge, we adopt GRPO as the base reinforcement learning framework, enabling domain knowledge to be integrated directly at the policy optimization level.
Our method consists of two key components.
First, we construct a \emph{domain-support sampling distribution} that explicitly encodes domain knowledge, and introduce a domain-aware constraint to align the policy sampling distribution with this domain support during reinforcement learning.
This constraint encourages domain-consistent behaviors such as invariance and symmetry.
Second, we quantify the degree of domain knowledge exhibited by each sampled output by measuring the divergence between the policy distribution and the domain-support distribution, and use this signal to reweight reinforcement learning advantages.
Together, these two mechanisms enable effective and stable injection of abstract domain knowledge into MLLMs.

\subsection{Preliminary}
\label{sec:preliminary}

GRPO~\cite{shao2024deepseekmath} is a value-free policy optimization method designed to improve training stability and sample efficiency in reinforcement learning. 
Derived from Proximal Policy Optimization (PPO)~\cite{schulman2017proximal}, GRPO avoids explicit value function estimation by adopting a group-based relative advantage formulation, which effectively regularizes policy updates while preserving sufficient optimization flexibility.
Let $\pi_{\theta}$ denote a policy parameterized by $\theta$. 
Given an input context $c$, GRPO first draws a group of $G$ candidate outputs $\{o_1, o_2, \dots, o_G\}$ from the previous policy $\pi_{\theta_{\mathrm{old}}}$. 
Each sampled output is then evaluated by predefined, verifiable reward functions, producing a set of scalar rewards $\{r_1, r_2, \dots, r_G\}$. 
Instead of relying on a learned value baseline, GRPO computes relative advantages within the sampled group by normalizing rewards as
\begin{equation}
\label{eq:ro}
A_i = 
\frac{r_i - \mathrm{mean}(\{r_1, r_2, \dots, r_G\})}
{\mathrm{std}(\{r_1, r_2, \dots, r_G\})}
\text{.}
\end{equation}

Using the resulting group-relative advantages $\{A_1, A_2, \dots, A_G\}$, the policy $\pi_{\theta}$ is optimized by maximizing the following objective:

\begin{equation}
\label{eq:grpo}
\footnotesize
\mathcal{L}(\theta) =
\mathbb{E}_{\{o_i\}_{i=1}^{G} \sim \pi_{\theta_{\mathrm{old}}}}
\left[
\frac{1}{G} \sum_{i=1}^{G}
\frac{\pi_{\theta}(o_i)}{\pi_{\theta_{\mathrm{old}}}(o_i)} A_i
- \beta \, \mathbb{D}_{\mathrm{KL}}
\big( \pi_{\theta} \,\Vert\, \pi_{\mathrm{ref}} \big)
\right]
\text{,}
\end{equation}
where $\mathbb{D}_{\mathrm{KL}}(\cdot \Vert \cdot)$ denotes the Kullback--Leibler divergence that constrains the updated policy $\pi_{\theta}$ to remain close to a reference policy $\pi_{\mathrm{ref}}$, and $\beta$ controls the strength of this regularization.

\subsection{Domain-aware Constraints}
We adopt Group Relative Policy Optimization (GRPO) as the base reinforcement learning framework and employ the reward functions proposed in Visual-RFT~\cite{Liu_2025_ICCV} to supervise policy optimization.
Given multimodal inputs consisting of prompt questions and domain-specific images, the model produces a policy sampling distribution $\pi_{\theta}$, while a reference model provides a reference distribution $\pi_{\mathrm{ref}}$.
In GRPO, the Kullback--Leibler (KL) divergence between $\pi_{\theta}$ and $\pi_{\mathrm{ref}}$ is used to regularize policy updates, ensuring that the optimized policy remains close to the reference model and thus preserves general prior knowledge.
Inspired by this mechanism, we further introduce a \emph{domain-aware constraint} to explicitly inject domain knowledge into the policy optimization process.

Specifically, we consider rotation invariance in remote sensing and symmetry consistency in medical imaging as representative forms of domain knowledge.
A straightforward approach would be to directly enforce invariance or consistency at the output or reward level. However, such explicit constraints are often brittle and task-dependent, and may fail to generalize across different prompts or output structures. Instead, we incorporate domain knowledge at the distribution level by regulating the policy sampling behavior under domain-specific transformations. To construct a domain-support sampling distribution, we apply domain-specific transformations, random rotations for remote sensing images and symmetric transformations for medical images, and feed the transformed inputs, together with the same prompt questions, into the policy model.
This process yields a domain-support sampling distribution denoted as $\pi_{\theta}^{D}$.
To enforce invariance and consistency, we minimize the KL divergence between the original policy distribution $\pi_{\theta}$ and the domain-support distribution $\pi_{\theta}^{D}$, which defines the domain loss:
\begin{equation}
    \mathcal{L}_{dom} = \mathbb{D}_{\mathrm{KL}}
    \big( \pi_{\theta}^{D} \,\Vert\, \pi_{\theta} \big).
\end{equation}
By constraining the policy sampling distribution to remain consistent under domain-specific transformations, this loss encourages the model to exhibit domain-invariant and domain-consistent behaviors.

\subsection{Domain-aware Reward Shaping}

Beyond domain-aware constraints, we further introduce \emph{domain-aware reward shaping} to provide fine-grained optimization signals that explicitly emphasize domain-consistent behaviors.
While the constraint term enforces distribution-level invariance globally, it treats all samples equally.
In contrast, domain-aware reward shaping allows the optimization process to adaptively focus on samples that better satisfy domain principles.

Inspired by group-based relative evaluation in GRPO, we quantify the degree of domain knowledge exhibited by each sampled output by measuring the divergence between the policy sampling distribution $\pi_{\theta}$ and the domain-support distribution $\pi_{\theta}^{D}$. We use KL divergence for constraints due to its asymmetry and stronger penalization of distributional mismatch, while JS divergence is adopted for advantage reweighting due to its boundedness and stability
Specifically, we compute the Jensen--Shannon (JS) divergence as
\begin{equation}
    \mathcal{D}_{i} = \mathbb{D}_{\mathrm{JS}}
    \big( \pi_{\theta}^{D} \,\Vert\, \pi_{\theta} \big),
\end{equation}
which yields a bounded and symmetric measure in $[0,1]$ that reflects how well a sample aligns with the domain-support distribution.
Using this divergence, we reweight the original GRPO advantages to obtain \emph{domain-aware advantages}.
Samples that are more consistent with domain knowledge (i.e., smaller divergence) are assigned larger advantages:
\begin{equation}
    A_i^{d} = (1 - \mathcal{D}_{i}) \cdot A_i .
\end{equation}

By incorporating the domain-aware advantages, the GRPO objective is modified as
\begin{align}
\footnotesize
\mathcal{L}(\theta)
&= \mathbb{E}_{\{o_i\}_{i=1}^{G} \sim \pi_{\theta_{\mathrm{old}}}}
\Bigg[
\frac{1}{G} \sum_{i=1}^{G}
\frac{\pi_{\theta}(o_i)}{\pi_{\theta_{\mathrm{old}}}(o_i)} A_i^{d}
\nonumber \\
&\qquad
- \beta \, \mathbb{D}_{\mathrm{KL}}
\big( \pi_{\theta} \,\Vert\, \pi_{\mathrm{ref}} \big)
- \mathbb{D}_{\mathrm{KL}}
\big( \pi_{\theta}^{D} \,\Vert\, \pi_{\theta} \big)
\Bigg].
\end{align}

More generally, the resulting \emph{domain-aware reinforcement learning objective} can be written as
\begin{align}
\footnotesize
\mathcal{L}_{\mathrm{DA}}(\theta)
&= \mathbb{E}_{\{o_i\}_{i=1}^{G} \sim \pi_{\theta_{\mathrm{old}}}}
\Bigg[
\frac{1}{G} \sum_{i=1}^{G}
\frac{\pi_{\theta}(o_i)}{\pi_{\theta_{\mathrm{old}}}(o_i)} A_i^{d}
\nonumber \\
&\qquad
- \beta \, \mathbb{D}_{\mathrm{KL}}
\big( \pi_{\theta} \,\Vert\, \pi_{\mathrm{ref}} \big)
- \mathcal{L}_{\mathrm{dom}}
\Bigg].
\end{align}
Although we instantiate  framework with rotation invariance and symmetry consistency, the proposed formulation is general and can incorporate arbitrary domain priors that can be expressed via transformations or distributional constraints.

\begin{table}[ht!]
\caption{
% \small
\textbf{Few-shot results on the remote sensing domain.} We conducted experiments on six remote sensing datasets. $\Delta$ denotes the performance benefits compared with SFT.}
\vspace{-2mm}
\label{tab2}
\begin{center}
\setlength{\tabcolsep}{4pt}
\scalebox{0.88}
{
\begin{tabular}{llcccccc}
\toprule
\multicolumn{1}{c}{\bf Models}&\multicolumn{1}{c}{\bf Average} &\rotatebox{90}{UCM} &\rotatebox{90}{AID} &\rotatebox{90}{RSICD} &\rotatebox{90}{WHURS19} &\rotatebox{90}{PatternNet} &\rotatebox{90}{NWPU}
\\
% \midrule
% \rowcolor[HTML]{F2F2F2} \multicolumn{8}{c}{\textbf{\textit{Qwen2-VL-2B}}} \\
\midrule
Qwen2.5-VL&16.0 &8.91	&18.6	&15.5	&29.3	&8.61	&14.9\\

\midrule
\multicolumn{8}{c}{\textbf{\textit{1-shot}}} \\
\midrule
SFT &16.8 &10.1	&19.8	&15.7	&30.3	&9.15 &15.6 \\
Visual-RFT &17.4 &10.3	&20.9	&16.1	&30.7	&9.61 &16.5 \\
\rowcolor[HTML]{DAEFF9} Ours &19.7  &16.1	&22.0	&17.1	&33.7	&10.8 &18.4 \\
$\Delta$ & \hgreen{+2.9} & \hgreen{+6.0} & \hgreen{+2.2} & \hgreen{+1.4} & \hgreen{+3.4} & \hgreen{+1.7} & \hgreen{+2.8} \\

\midrule
\multicolumn{8}{c}{\textbf{\textit{2-shot}}} \\
\midrule
SFT &18.3 &14.9	&20.6	&16.0	&31.2	&10.4 &16.6 \\
Visual-RFT &19.1 &15.7	&21.8	&16.2	&31.7	&11.9	&17.2 \\
\rowcolor[HTML]{DAEFF9} Ours & 20.9 &17.1	&23.8	&19.6	&33.6	&12.6 &18.7 \\
$\Delta$ & \hgreen{+2.6} & \hgreen{+2.2} & \hgreen{+3.2} & \hgreen{+3.6} & \hgreen{+2.4} & \hgreen{+1.2} & \hgreen{+2.1} \\

\midrule
\multicolumn{8}{c}{\textbf{\textit{4-shot}}} \\
\midrule
SFT &20.3 &17.0	&21.8	&19.8	&32.3	&12.0 &18.9 \\
Visual-RFT &21.9 &18.4	&22.4	&23.5	&32.6	&12.5		&21.8 \\
\rowcolor[HTML]{DAEFF9} Ours &25.0  &20.1	&24.4	&25.8 	&38.8	&17.5		&23.5 \\
$\Delta$ & \hgreen{+4.7} & \hgreen{+3.1} & \hgreen{+2.6} & \hgreen{+6.0} & \hgreen{+6.5} & \hgreen{+5.5} & \hgreen{+4.6} \\

\midrule
\multicolumn{8}{c}{\textbf{\textit{8-shot}}} \\
\midrule
SFT &25.5 &21.8	&29.9	&22.6	&39.5	&15.4 &23.6 \\
Visual-RFT &28.5 &23.6	&32.9	&25.6	&40.2	&18.2  &30.5 \\
\rowcolor[HTML]{DAEFF9} Ours &32.3   &28.8	&38.9	&26.4	&44.4	&19.1	&36.4\\
$\Delta$ & \hgreen{+7.8} & \hgreen{+7.0} & \hgreen{+9.0} & \hgreen{+3.8} & \hgreen{+4.9} & \hgreen{+3.7} & \hgreen{+12.8} \\

\bottomrule
\end{tabular}
}
\vspace{-7mm}
\end{center}
\end{table}

\begin{table}[t!]
\caption{
% \small
\textbf{Few-shot results on the medical imaging domain.} We conducted experiments on five medical imaging datasets. $\Delta$ denotes the performance benefits compared with SFT.}
\vspace{-2mm}
\label{tab3}
\begin{center}
\setlength{\tabcolsep}{4pt}
\scalebox{1.0}
{
\begin{tabular}{llccccc}
\toprule
\multicolumn{1}{c}{\bf Models}&\multicolumn{1}{c}{\bf Average} &\rotatebox{90}{OrganAMNIST} &\rotatebox{90}{BloodMNIST} &\rotatebox{90}{DermaMNIST} &\rotatebox{90}{PathMNIST} &\rotatebox{90}{TissueMNIST}
\\
% \midrule
% \rowcolor[HTML]{F2F2F2} \multicolumn{8}{c}{\textbf{\textit{Qwen2-VL-2B}}} \\
\midrule
Qwen2.5-VL&20.7 & 8.70	&14.8	&21.7	&29.0	&29.1  \\

\midrule
\multicolumn{7}{c}{\textbf{\textit{1-shot}}} \\
\midrule
SFT &20.9 &9.10	&15.2	&21.3	&29.8	&29.5 \\
Visual-RFT &21.0 &9.39	&15.2	&21.1	&29.9	&29.5 \\
\rowcolor[HTML]{DAEFF9} Ours &21.5 &9.85	&15.6	&21.8	&30.6	&29.7  \\
$\Delta$ & \hgreen{+0.6} & \hgreen{+0.75} & \hgreen{+0.4} & \hgreen{+0.5} & \hgreen{+0.8} & \hgreen{+0.2}  \\

\midrule
\multicolumn{7}{c}{\textbf{\textit{2-shot}}} \\
\midrule
SFT &21.4 &9.76	&15.5	&21.8	&30.1 &29.8 \\
Visual-RFT &21.6 &9.81	&15.6	&22.3	&30.3	&29.8\\
\rowcolor[HTML]{DAEFF9} Ours &23.2 &12.4	&16.1	&23.5	&33.2	&30.9 \\
$\Delta$ & \hgreen{+1.8} & \hgreen{+2.6} & \hgreen{+0.6} & \hgreen{+1.7} & \hgreen{+3.1} & \hgreen{+1.1} \\

\midrule
\multicolumn{7}{c}{\textbf{\textit{4-shot}}} \\
\midrule
SFT &22.9 &10.0	&16.1	&27.3	&30.6	&30.3  \\
Visual-RFT &23.0 &10.1	&16.0	&27.6	&30.9	&30.6 \\
\rowcolor[HTML]{DAEFF9} Ours &25.3  &13.2	&17.2	&32.8 	&31.4	&31.7\\
$\Delta$ & \hgreen{+2.4} & \hgreen{+3.2} & \hgreen{+1.1} & \hgreen{+5.5} & \hgreen{+0.8} & \hgreen{+1.4}\\

\midrule
\multicolumn{7}{c}{\textbf{\textit{8-shot}}} \\
\midrule
SFT &24.1 &10.7	&17.1	&28.9	&32.3	&31.4\\
Visual-RFT &24.4 &10.8	&17.4	&29.9	&32.6	&31.5\\
\rowcolor[HTML]{DAEFF9} Ours &26.8  &13.3	&18.9	&33.1	&35.4	&33.1\\
$\Delta$ & \hgreen{+2.7} & \hgreen{+2.5} & \hgreen{+1.8} & \hgreen{+4.2} & \hgreen{+3.1} & \hgreen{+1.7} \\

\bottomrule
\end{tabular}
}
\vspace{-6mm}
\end{center}
\end{table}

\begin{table}[htbp]
\caption{
% \small
\textbf{Few-shot grounding results on DIOR dataset of 6 representative categories.} We conducted 4-shot and 8-shot experiments. $\Delta$ denotes the performance benefits compared with SFT. The first column denotes the average mAP.}
\vspace{-2mm}
\label{tab4}
\begin{center}
\setlength{\tabcolsep}{2pt}
\scalebox{1.0}
{
\begin{tabular}{llcccccc}
\toprule
\multicolumn{1}{c}{\bf Models} &\multicolumn{1}{c}{\bf mAP} &\rotatebox{90}{\bf Airplane} &\rotatebox{90}{\bf Bridge} &\rotatebox{90}{\bf Harbor} &\rotatebox{90}{\bf Stadium}&\rotatebox{90}{\bf Storage Tank}&\rotatebox{90}{\bf Ship}
\\
\midrule
Qwen2.5-VL &11.3 & 15.6 & 5.71 & 13.2 & 19.8 & 8.66 & 4.60\\
\midrule
\multicolumn{8}{c}{\textbf{\textit{4-shot}}} \\
\midrule

SFT & 17.8 & 22.2 & 15.4 & 15.8 & 23.8 & 13.6 & 15.9\\
Visual-RFT & 19.6 & 25.3 & 17.7 & 16.8 & 24.2 & 15.8 & 17.9 \\
\rowcolor[HTML]{DAEFF9}  Ours & 21.0 & 27.6 & 18.6 & 17.9 & 25.8 & 16.7 & 19.5 \\
$\Delta$ & \hgreen{+3.2} & \hgreen{+5.4} & \hgreen{+3.2} & \hgreen{+2.1} & \hgreen{+2.0}& \hgreen{+3.1}& \hgreen{+3.6}\\

\midrule
\multicolumn{8}{c}{\textbf{\textit{8-shot}}} \\
\midrule

SFT &22.1  & 27.1 & 17.9 & 26.7 & 25.1 & 16.9 & 18.7 \\
Visual-RFT &22.7 & 27.3 & 18.8 & 27.9 & 25.9 & 17.1 & 19.2 \\
\rowcolor[HTML]{DAEFF9}  Ours &24.6 &29.9 & 20.6 &29.6 & 27.8 & 18.4 & 21.3\\
$\Delta$ & \hgreen{+2.5} & \hgreen{+2.8} & \hgreen{+2.7} & \hgreen{+2.9} & \hgreen{+2.7}& \hgreen{+2.5}& \hgreen{+2.6}\\
\bottomrule
\end{tabular}
}
\vspace{-6mm}
\end{center}
\end{table}

\section{Experiments}
\subsection{Experimental Setups}
To evaluate our method under realistic specialized-domain scenarios, we conduct few-shot learning experiments in both remote sensing and medical imaging domains.
For the remote sensing domain, we consider six widely used benchmark datasets: UCM~\cite{yang2010bag}, AID~\cite{xia2017aid}, RSICD~\cite{lu2017exploring}, WHURS19~\cite{Dai2011WHURS19}, PatternNet~\cite{zhou2018patternnet}, and NWPU~\cite{cheng2017remote}.
For the medical imaging domain, we adopt datasets from MedMNIST v2~\cite{medmnistv2}, including OrganAMNIST, BloodMNIST, DermaMNIST, PathMNIST, and TissueMNIST.

We conducted experiments under 1-shot, 2-shot, 4-shot, and 8-shot settings, where the number of shots denotes the number of annotated samples per category. We adopt the reward functions in Visual-RFT~\cite{Liu_2025_ICCV} to reward the policy model. To further assess the effectiveness of our approach in multimodal reasoning, we additionally conduct few-shot grounding experiments on the DIOR~\cite{li2020object} dataset. All experiments are conducted using Qwen2.5-VL as the base MLLM. Training is performed with a batch size of 4 on 2 or 4 NVIDIA A100 GPUs (80GB). The model is trained for 2 epochs in the 1-shot and 2-shot settings, and for 4 epochs in the 4-shot and 8-shot settings. We use the Adam optimizer with a learning rate of $5 \times 10^{-5}$. Unless otherwise specified, all experiments use the same prompt templates and evaluation protocols.

\subsection{Quantitative Results}
\paragraph{Remote sensing domain.}
Table~\ref{tab2} reports the few-shot recognition results on six remote sensing datasets using Qwen2.5-VL-3B.Our method consistently outperforms both SFT and Visual-RFT across all shot settings and all datasets.
Notably, the performance gains become more pronounced as the number of shots increases, indicating that domain-aware reinforcement fine-tuning can more effectively leverage limited supervision.
On average, our approach improves over SFT by \textbf{+2.9}, \textbf{+2.6}, \textbf{+4.7}, and \textbf{+7.8} points under the 1-shot, 2-shot, 4-shot, and 8-shot settings, respectively.
Substantial improvements are observed on datasets that exhibit strong rotational characteristics, such as WHURS19, PatternNet, and NWPU, where our method achieves gains of up to \textbf{+6.5}, \textbf{+5.5}, and \textbf{+12.8} points.
These results validate the effectiveness of injecting domain knowledge at the policy optimization level.

\paragraph{Medical imaging domain.}
Table~\ref{tab3} summarizes the few-shot recognition results on five medical imaging benchmarks using Qwen2.5-VL-3B.
Compared with remote sensing, the improvements in the 1-shot setting are relatively modest, which is expected due to the higher visual homogeneity and label ambiguity in medical images.
Nevertheless, our method consistently outperforms both SFT and Visual-RFT across all shot settings, achieving average gains over SFT of \textbf{+0.6}, \textbf{+1.8}, \textbf{+2.4}, and \textbf{+2.7} points from 1-shot to 8-shot.
The advantages become more evident as the number of shots increases, particularly on datasets with strong anatomical symmetry such as OrganAMNIST and PathMNIST, where our method yields gains of up to \textbf{+3.2} and \textbf{+5.5} points.
These results demonstrate that enforcing symmetry consistency at the distribution level effectively injects structural medical priors and improves few-shot generalization.

\paragraph{Few-shot grounding.}
Following the experimental protocol of Visual-RFT~\cite{Liu_2025_ICCV}, we randomly sample six representative categories from the DIOR dataset and evaluate few-shot grounding performance using the larger Qwen2.5-VL-7B model, as reported in Table~\ref{tab4}.
Our method consistently outperforms both SFT and Visual-RFT under the 4-shot and 8-shot settings.
In particular, it achieves average mAP improvements of \textbf{+3.2} and \textbf{+2.5} over SFT in the 4-shot and 8-shot scenarios, respectively.
Performance gains are observed across all six object categories, including Airplane, Bridge, and Ship, demonstrating more accurate spatial localization and region-level reasoning.
These results indicate that domain-aware reinforcement fine-tuning not only improves classification performance but also effectively enhances multimodal grounding capabilities when applied to larger-capacity models.

\begin{figure*}[t]
    \begin{center}
    %\framebox[4.0in]{$\;$}
    \includegraphics[width=0.95\linewidth]{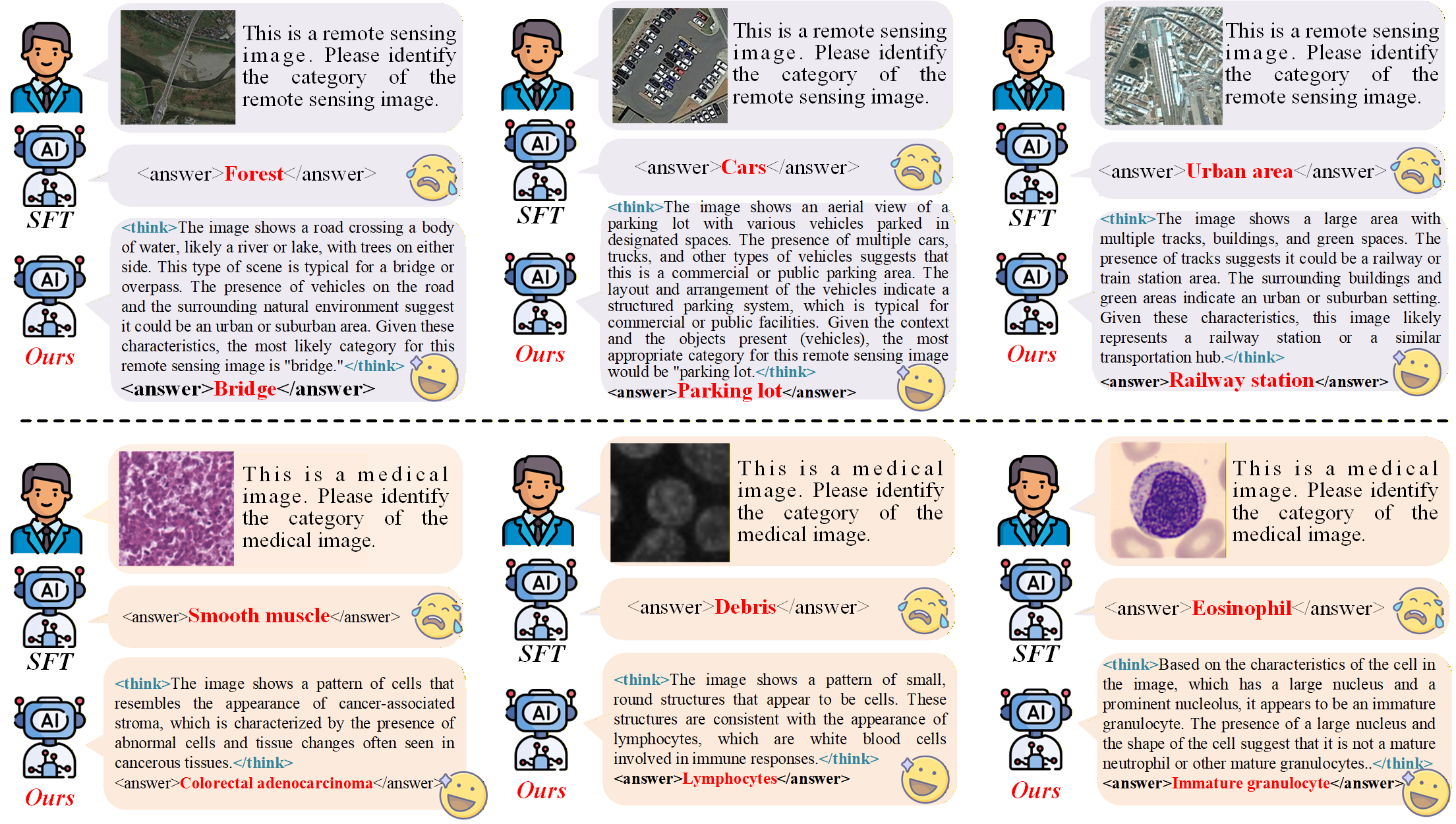}
    \end{center}
    \vspace{-2mm}
\caption{\small 
\textbf{Qualitative results on remote sensing and medical imaging tasks.}
Compared with directly supervised fine-tuning~(SFT), our method yields accurate and fine-grained categories by grounding predictions in discriminative visual patterns and domain-specific cues. First Row: Remote Sensing. Second Row: Medical Imaging.}
    \label{fig3}
\vspace{-4mm}
\end{figure*}

\subsection{Qualitative Results}
Figure~\ref{fig3} presents the qualitative comparisons between SFT and our method across both remote sensing and medical imaging domains. Particularly, the first row denotes samples in the remote sensing domain, while the second row denotes results in medical imaging domain. For remote sensing scenes, SFT often produces coarse or semantically incorrect predictions, such as confusing bridges with forests, parking lots with generic vehicle-related categories, or railway stations with urban areas.
In contrast, our method correctly identifies fine-grained scene categories by grounding predictions in discriminative spatial structures and object arrangements, e.g., linear bridge spans over water, dense vehicle layouts in parking lots, and track-centered configurations in railway stations.
Similar improvements are observed in the medical imaging domain.
While SFT tends to generate superficial or noisy predictions (e.g., confusing debris with lymphocytes or misclassifying tissue types), our approach produces clinically meaningful categories such as colorectal adenocarcinoma, lymphocytes, and immature granulocytes.
These predictions are supported by coherent visual reasoning that aligns cellular morphology with domain-specific semantics.
Overall, the results demonstrate that domain-aware reinforcement fine-tuning effectively enhances cross-domain visual reasoning and semantic grounding. Our method not only improves answer correctness but also yields more interpretable and visually consistent predictions, highlighting its robustness in both complex remote sensing scenes and fine-grained medical imagery.

\begin{table}[t]
\centering
\caption{Ablation study of domain-aware reinforcement learning components on the UCM dataset.
DC denotes domain-aware constraint and DR denotes domain-aware reward shaping.}
\label{tab:ablation_components}
\vspace{-2mm}
% \small
\begin{tabular}{cc|cccc}
\midrule
\hline
DC & DR & 1-shot & 2-shot & 4-shot & 8-shot \\
\hline
$\times$ & $\times$ & 10.3 & 15.7 & 18.4 & 23.6 \\
$\checkmark$ & $\times$ & 14.5 & 16.8 & 19.1 & 26.5 \\
$\times$ & $\checkmark$ & 15.4 & 16.6 & 19.8 & 27.7 \\
\rowcolor[HTML]{DAEFF9}
$\checkmark$ & $\checkmark$ & \textbf{16.1} & \textbf{17.1} & \textbf{20.1} & \textbf{28.8} \\
\hline
\end{tabular}
\vspace{-4mm}
\end{table}

\subsection{Ablation Study}
\paragraph{Effect of domain-aware reinforcement learning components.}
Table~\ref{tab:ablation_components} reports the ablation results of the proposed domain-aware reinforcement learning components on the UCM dataset under different few-shot settings. Starting from the baseline without any domain-aware design, introducing either the domain-aware constraint (DC) or the domain-aware reward shaping (DR) consistently improves performance across all shot numbers, demonstrating that both components are individually effective. In particular, DC yields more pronounced gains in extremely low-shot settings, indicating its effectiveness in constraining the policy toward domain-consistent behaviors, while DR provides larger improvements in higher-shot regimes by delivering more informative and stable reward signals. When both DC and DR are jointly applied, the model achieves the best performance in all settings, with clear and consistent margins over using either component alone. These results validate the complementary nature of DC and DR, and confirm that their combination is crucial for fully exploiting domain knowledge in reinforcement fine-tuning.

\begin{table}[t]
\centering
\caption{Comparison with direct data augmentation on the UCM dataset.
DA denotes standard data augmentation applied during training.VRFT denotes Visual-RFT~\cite{Liu_2025_ICCV}.}
\label{tab:ablation_augmentation}
\vspace{-2mm}
\scalebox{0.9}{
\begin{tabular}{lcccc}
\toprule
% \hline
Method &  1-shot & 2-shot & 4-shot & 8-shot \\
\midrule
SFT & 10.1 & 14.9 &17.0 &21.8 \\
SFT + DA & 10.2 & 14.8 & 17.1 & 21.6\\
VRFT & 10.3 & 15.7 & 18.4 & 23.6 \\
VRFT + DA & 10.4 & 15.8 & 18.3 & 23.7\\
\rowcolor[HTML]{DAEFF9}
Ours & \textbf{16.1} & \textbf{17.1} & \textbf{20.1} & \textbf{28.8}  \\
\bottomrule
\end{tabular}
}
\vspace{-4mm}
\end{table}

\paragraph{Comparison with Direct Data Augmentation.}
Table~\ref{tab:ablation_augmentation} compares our method with standard data augmentation (DA) strategies applied during training on the UCM dataset. As shown, incorporating conventional data augmentation into either SFT or Visual-RFT leads to only marginal performance changes, and in some cases even results in slight degradation, particularly under low-shot settings. This suggests that naive data-level transformations are insufficient to effectively capture domain-specific invariances and may introduce additional noise when supervision is extremely limited. In contrast, our domain-aware reinforcement fine-tuning significantly outperforms all baselines by a large margin across all shot numbers. Notably, the performance gap becomes increasingly pronounced as the number of shots increases, highlighting that explicitly injecting domain knowledge at the policy optimization level is substantially more effective than relying on direct data augmentation. These results further confirm that domain-aware RL provides a more principled and robust mechanism for exploiting structural priors than conventional augmentation-based approaches.

\begin{table}[t]
\centering
\caption{Ablation study comparing output-level constraints and distribution-level constraints on the UCM dataset.OC denotes output-level constraint, DC denotes domain-aware
constraint and DR denotes domain-aware reward shaping.}
\label{tab:ablation_output_vs_distribution}
\small
\begin{tabular}{lcccc}
\toprule
Method &   1-shot & 2-shot & 4-shot & 8-shot \\
\midrule
Baseline & 10.3 & 15.7 & 18.4 & 23.6\\
Baseline + OC & 11.5 & 15.9 & 18.6 & 23.5\\
Baseline + DC & 14.5 & 16.8 & 19.1 & 26.5\\
\rowcolor[HTML]{DAEFF9}
DC + DR (Ours) & \textbf{16.1} & \textbf{17.1} & \textbf{20.1} & \textbf{28.8} \\
\bottomrule
\end{tabular}
\vspace{-3mm}
\end{table}

\paragraph{Output-level vs. Distribution-level Constraints.}
Table~\ref{tab:ablation_output_vs_distribution} investigates the effectiveness of enforcing domain knowledge at different levels, comparing output-level constraints (OC) with distribution-level constraints incorporated into reinforcement learning. Applying output-level constraints on top of the baseline yields only marginal improvements, and the gains are inconsistent across shot settings, indicating limited capability in capturing domain-specific invariances. In contrast, introducing domain-aware constraints at the distribution level (DC) leads to substantially larger and more stable performance gains across all shot numbers, demonstrating that enforcing consistency over model behaviors is more effective than directly regularizing final predictions. Furthermore, combining distribution-level constraints with domain-aware reward shaping (DC + DR) achieves the best performance by a clear margin, highlighting the complementary roles of constraint-based regularization and reward-driven optimization. These results confirm that injecting domain knowledge at the policy and distribution level is crucial for robust few-shot generalization, while output-level heuristics alone are insufficient.

\begin{table}[t]
\centering
\caption{Effect of divergence choices for domain-aware constraint (DC) and domain-aware reward shaping (DR) on the UCM dataset. Div. denotes Divergence.}
\label{tab:ablation_divergence}
\vspace{-2mm}
\scalebox{0.9}{
\begin{tabular}{cc|cccc}
\midrule
\hline
DC Div. & DR Div. &1-shot & 2-shot & 4-shot & 8-shot\\
\hline
KL & KL & 14.8 & 16.5 & 18.7 & 26.5 \\
JS & JS & 15.7 & 16.8 & 19.5 & 27.6 \\
JS & KL & 13.9 & 16.1 & 17.6 & 25.3 \\
\rowcolor[HTML]{DAEFF9}
KL(Ours)  & JS (Ours) & \textbf{16.1} & \textbf{17.1} & \textbf{20.1} & \textbf{28.8} \\
\hline
\end{tabular}
}
\vspace{-4mm}
\end{table}

\paragraph{Divergence Choices for Varying Components.}
Table~\ref{tab:ablation_divergence} analyzes the impact of different divergence functions used in domain-aware constraint (DC) and domain-aware reward shaping (DR). When the same divergence is applied to both components (KL–KL or JS–JS), performance improves over the baseline but remains suboptimal, suggesting that DC and DR play distinct roles and benefit from different divergence characteristics. Using JS divergence for DC provides more stable improvements than KL, as JS is symmetric and bounded, making it better suited for enforcing distribution-level consistency. Conversely, employing KL divergence for DR leads to inferior performance, indicating that asymmetric penalties in reward shaping may destabilize policy optimization. Our final design, which applies KL divergence for DC and JS divergence for DR, achieves the best results across all shot settings. This complementary combination effectively balances strict distribution alignment and stable reward optimization, highlighting the importance of carefully selecting divergence functions for different components in domain-aware reinforcement learning.

\section{Conclusion}
We propose a domain-aware reinforcement fine-tuning framework that injects abstract domain knowledge into multimodal large language models at the policy distribution level.
By combining domain-aware constraints with domain-aware reward shaping, our method enforces domain-knowledge consistent behaviors directly during reinforcement learning.
Extensive experiments across remote sensing and medical imaging benchmarks demonstrate consistent improvements under few-shot settings, and ablation studies confirm the complementary roles of the proposed components. These results suggest that domain-aware reinforcement learning is an effective and general approach for adapting multimodal foundation models to specialized domains with limited supervision.
We believe this framework provides a promising direction for systematically integrating domain priors into multimodal models beyond the specific tasks studied in this work.


% Authors are kindly asked to make their submissions as accessible as possible
% for everyone including people with disabilities and sensory or neurological
% differences. Tips of how to achieve this and what to pay attention to will be
% provided on the conference website \url{http://icml.cc/}.

% \section*{Software and Data}

% If a paper is accepted, we strongly encourage the publication of software and
% data with the camera-ready version of the paper whenever appropriate. This can
% be done by including a URL in the camera-ready copy. However, \textbf{do not}
% include URLs that reveal your institution or identity in your submission for
% review. Instead, provide an anonymous URL or upload the material as
% ``Supplementary Material'' into the OpenReview reviewing system. Note that
% reviewers are not required to look at this material when writing their review.

% % Acknowledgements should only appear in the accepted version.
% \section*{Acknowledgements}

% \textbf{Do not} include acknowledgements in the initial version of the paper

% If a paper is accepted, the final camera-ready version can (and usually should)
% include acknowledgements.  Such acknowledgements should be placed at the end of
% the section, in an unnumbered section that does not count towards the paper
% page limit. Typically, this will include thanks to reviewers who gave useful
% comments, to colleagues who contributed to the ideas, and to funding agencies
% and corporate sponsors that provided financial support.

\section*{Impact Statement}
This paper presents work whose goal is to advance the field of machine learning. There are many potential societal consequences of our work, none of which we feel must be specifically highlighted here.

% Authors are \textbf{required} to include a statement of the potential broader
% impact of their work, including its ethical aspects and future societal
% consequences. This statement should be in an unnumbered section at the end of
% the paper (co-located with Acknowledgements -- the two may appear in either
% order, but both must be before References), and does not count toward the paper
% page limit. In many cases, where the ethical impacts and expected societal
% implications are those that are well established when advancing the field of
% Machine Learning, substantial discussion is not required, and a simple
% statement such as the following will suffice:

% ``This paper presents work whose goal is to advance the field of Machine
% Learning. There are many potential societal consequences of our work, none
% which we feel must be specifically highlighted here.''

% The above statement can be used verbatim in such cases, but we encourage
% authors to think about whether there is content which does warrant further
% discussion, as this statement will be apparent if the paper is later flagged
% for ethics review.

% % In the unusual situation where you want a paper to appear in the
% % references without citing it in the main text, use \nocite

\bibliography{example_paper}

@article{achiam2023gpt,
  title={Gpt-4 technical report},
  author={Achiam, Josh and Adler, Steven and Agarwal, Sandhini and Ahmad, Lama and Akkaya, Ilge and Aleman, Florencia Leoni and Almeida, Diogo and Altenschmidt, Janko and Altman, Sam and Anadkat, Shyamal and others},
  journal={arXiv:2303.08774},
  year={2023}
}

@article{bai2025qwen2,
  title={Qwen2. 5-vl technical report},
  author={Bai, Shuai and Chen, Keqin and Liu, Xuejing and Wang, Jialin and Ge, Wenbin and Song, Sibo and Dang, Kai and Wang, Peng and Wang, Shijie and Tang, Jun and others},
  journal={arXiv:2502.13923},
  year={2025}
}

@article{lu2025ovis2,
  title={Ovis2. 5 technical report},
  author={Lu, Shiyin and Li, Yang and Xia, Yu and Hu, Yuwei and Zhao, Shanshan and Ma, Yanqing and Wei, Zhichao and Li, Yinglun and Duan, Lunhao and Zhao, Jianshan and others},
  journal={arXiv:2508.11737},
  year={2025}
}

@inproceedings{jiang2025corvid,
  title={Corvid: Improving multimodal large language models towards chain-of-thought reasoning},
  author={Jiang, Jingjing and Ma, Chao and Song, Xurui and Zhang, Hanwang and Luo, Jun},
  booktitle={Proceedings of the IEEE/CVF International Conference on Computer Vision},
  pages={3034--3046},
  year={2025}
}

@inproceedings{wang2025triplets,
  title={Triplets better than pairs: Towards stable and effective self-play fine-tuning for LLMs},
  author={Wang, Yibo and Sun, Hai-Long and Huzhang, Guangda and Chen, Qing-Guo and Xu, Zhao and Luo, Weihua and Zhang, Kaifu and Zhang, Lijun},
  booktitle={The Thirty-ninth Annual Conference on Neural Information Processing Systems},
  year={2025}
}

@inproceedings{sunparrot,
  title={Parrot: Multilingual Visual Instruction Tuning},
  author={Sun, Hai-Long and Zhou, Da-Wei and Li, Yang and Lu, Shiyin and Yi, Chao and Chen, Qing-Guo and Xu, Zhao and Luo, Weihua and Zhang, Kaifu and Zhan, De-Chuan and others},
  booktitle={Forty-second International Conference on Machine Learning},
  year={2025}
}

@article{xu2025probing,
  title={Probing Scientific General Intelligence of LLMs with Scientist-Aligned Workflows},
  author={Xu, Wanghan and Zhou, Yuhao and Zhou, Yifan and Cao, Qinglong and Li, Shuo and Bu, Jia and Liu, Bo and Chen, Yixin and He, Xuming and Zhao, Xiangyu and others},
  journal={arXiv:2512.16969},
  year={2025}
}

@article{jiang2025multimodal,
  title={Multimodal Tabular Reasoning with Privileged Structured Information},
  author={Jiang, Jun-Peng and Xia, Yu and Sun, Hai-Long and Lu, Shiyin and Chen, Qing-Guo and Luo, Weihua and Zhang, Kaifu and Zhan, De-Chuan and Ye, Han-Jia},
  journal={arXiv:2506.04088},
  year={2025}
}

@inproceedings{li2023blip,
  title={Blip-2: Bootstrapping language-image pre-training with frozen image encoders and large language models},
  author={Li, Junnan and Li, Dongxu and Savarese, Silvio and Hoi, Steven},
  booktitle={International conference on machine learning},
  pages={19730--19742},
  year={2023},
  organization={PMLR}
}

@article{guo2025deepseek,
  title={DeepSeek-R1 incentivizes reasoning in LLMs through reinforcement learning},
  author={Guo, Daya and Yang, Dejian and Zhang, Haowei and Song, Junxiao and Wang, Peiyi and Zhu, Qihao and Xu, Runxin and Zhang, Ruoyu and Ma, Shirong and Bi, Xiao and others},
  journal={Nature},
  volume={645},
  number={8081},
  pages={633--638},
  year={2025},
  publisher={Nature Publishing Group UK London}
}

@inproceedings{zhang2024large,
  title={Large-scale reinforcement learning for diffusion models},
  author={Zhang, Yinan and Tzeng, Eric and Du, Yilun and Kislyuk, Dmitry},
  booktitle={European Conference on Computer Vision},
  pages={1--17},
  year={2024},
  organization={Springer}
}

@article{wang2024rl,
  title={Rl-vlm-f: Reinforcement learning from vision language foundation model feedback},
  author={Wang, Yufei and Sun, Zhanyi and Zhang, Jesse and Xian, Zhou and Biyik, Erdem and Held, David and Erickson, Zackory},
  journal={arXiv:2402.03681},
  year={2024}
}

@article{yue2025does,
  title={Does reinforcement learning really incentivize reasoning capacity in llms beyond the base model?},
  author={Yue, Yang and Chen, Zhiqi and Lu, Rui and Zhao, Andrew and Wang, Zhaokai and Song, Shiji and Huang, Gao},
  journal={arXiv:2504.13837},
  year={2025}
}

@inproceedings{dong2024abilities,
  title={How abilities in large language models are affected by supervised fine-tuning data composition},
  author={Dong, Guanting and Yuan, Hongyi and Lu, Keming and Li, Chengpeng and Xue, Mingfeng and Liu, Dayiheng and Wang, Wei and Yuan, Zheng and Zhou, Chang and Zhou, Jingren},
  booktitle={Proceedings of the 62nd Annual Meeting of the Association for Computational Linguistics (Volume 1: Long Papers)},
  pages={177--198},
  year={2024}
}

@article{zhang2025domain,
  title={Domain-specific large language models for fault diagnosis of heating, ventilation, and air conditioning systems by labeled-data-supervised fine-tuning},
  author={Zhang, Jian and Zhang, Chaobo and Lu, Jie and Zhao, Yang},
  journal={Applied Energy},
  volume={377},
  pages={124378},
  year={2025},
  publisher={Elsevier}
}

@article{yuan2024self,
  title={Self-play fine-tuning of diffusion models for text-to-image generation},
  author={Yuan, Huizhuo and Chen, Zixiang and Ji, Kaixuan and Gu, Quanquan},
  journal={Advances in Neural Information Processing Systems},
  volume={37},
  pages={73366--73398},
  year={2024}
}

@inproceedings{li2024seed,
  title={Seed-bench: Benchmarking multimodal large language models},
  author={Li, Bohao and Ge, Yuying and Ge, Yixiao and Wang, Guangzhi and Wang, Rui and Zhang, Ruimao and Shan, Ying},
  booktitle={Proceedings of the IEEE/CVF Conference on Computer Vision and Pattern Recognition},
  pages={13299--13308},
  year={2024}
}

@inproceedings{huang2024smartedit,
  title={Smartedit: Exploring complex instruction-based image editing with multimodal large language models},
  author={Huang, Yuzhou and Xie, Liangbin and Wang, Xintao and Yuan, Ziyang and Cun, Xiaodong and Ge, Yixiao and Zhou, Jiantao and Dong, Chao and Huang, Rui and Zhang, Ruimao and others},
  booktitle={Proceedings of the IEEE/CVF Conference on Computer Vision and Pattern Recognition},
  pages={8362--8371},
  year={2024}
}

@article{zang2025contextual,
  title={Contextual object detection with multimodal large language models},
  author={Zang, Yuhang and Li, Wei and Han, Jun and Zhou, Kaiyang and Loy, Chen Change},
  journal={International Journal of Computer Vision},
  volume={133},
  number={2},
  pages={825--843},
  year={2025},
  publisher={Springer}
}

@inproceedings{luo2025mmevol,
  title={Mmevol: Empowering multimodal large language models with evol-instruct},
  author={Luo, Run and Zhang, Haonan and Chen, Longze and Lin, Ting-En and Liu, Xiong and Wu, Yuchuan and Yang, Min and Li, Yongbin and Wang, Minzheng and Zeng, Pengpeng and others},
  booktitle={Findings of the Association for Computational Linguistics: ACL 2025},
  pages={19655--19682},
  year={2025}
}

@inproceedings{wang2025divide,
  title={Divide, conquer and combine: A training-free framework for high-resolution image perception in multimodal large language models},
  author={Wang, Wenbin and Ding, Liang and Zeng, Minyan and Zhou, Xiabin and Shen, Li and Luo, Yong and Yu, Wei and Tao, Dacheng},
  booktitle={Proceedings of the AAAI Conference on Artificial Intelligence},
  volume={39},
  number={8},
  pages={7907--7915},
  year={2025}
}

@article{huang2024large,
  title={From large language models to large multimodal models: A literature review},
  author={Huang, Dawei and Yan, Chuan and Li, Qing and Peng, Xiaojiang},
  journal={Applied Sciences},
  volume={14},
  number={12},
  pages={5068},
  year={2024},
  publisher={MDPI}
}

@inproceedings{han2024multimodal,
  title={Multimodal large language models and tunings: Vision, language, sensors, audio, and beyond},
  author={Han, Soyeon Caren and Cao, Feiqi and Poon, Josiah and Navigli, Roberto},
  booktitle={Proceedings of the 32nd ACM International Conference on Multimedia},
  pages={11294--11295},
  year={2024}
}

@inproceedings{luo2025mono,
  title={Mono-internvl: Pushing the boundaries of monolithic multimodal large language models with endogenous visual pre-training},
  author={Luo, Gen and Yang, Xue and Dou, Wenhan and Wang, Zhaokai and Liu, Jiawen and Dai, Jifeng and Qiao, Yu and Zhu, Xizhou},
  booktitle={Proceedings of the Computer Vision and Pattern Recognition Conference},
  pages={24960--24971},
  year={2025}
}

@inproceedings{lin2024vila,
  title={Vila: On pre-training for visual language models},
  author={Lin, Ji and Yin, Hongxu and Ping, Wei and Molchanov, Pavlo and Shoeybi, Mohammad and Han, Song},
  booktitle={Proceedings of the IEEE/CVF conference on computer vision and pattern recognition},
  pages={26689--26699},
  year={2024}
}

@article{fan2024pre,
  title={On pre-training of multimodal language models customized for chart understanding},
  author={Fan, Wan-Cyuan and Chen, Yen-Chun and Liu, Mengchen and Yuan, Lu and Sigal, Leonid},
  journal={arXiv:2407.14506},
  year={2024}
}

@article{cheng2024domain,
  title={On Domain-Adaptive Post-Training for Multimodal Large Language Models},
  author={Cheng, Daixuan and Huang, Shaohan and Zhu, Ziyu and Zhang, Xintong and Zhao, Wayne Xin and Luan, Zhongzhi and Dai, Bo and Zhang, Zhenliang},
  journal={arXiv:2411.19930},
  year={2024}
}

@article{wang2024q,
  title={Q-vlm: Post-training quantization for large vision-language models},
  author={Wang, Changyuan and Wang, Ziwei and Xu, Xiuwei and Tang, Yansong and Zhou, Jie and Lu, Jiwen},
  journal={Advances in Neural Information Processing Systems},
  volume={37},
  pages={114553--114573},
  year={2024}
}

@inproceedings{cheng2025domain,
  title={On domain-adaptive post-training for multimodal large language models},
  author={Cheng, Daixuan and Huang, Shaohan and Zhu, Ziyu and Zhang, Xintong and Zhao, Wayne Xin and Luan, Zhongzhi and Dai, Bo and Zhang, Zhenliang},
  booktitle={Findings of the Association for Computational Linguistics: EMNLP 2025},
  pages={274--296},
  year={2025}
}

@article{liu2024deepseek,
  title={Deepseek-v3 technical report},
  author={Liu, Aixin and Feng, Bei and Xue, Bing and Wang, Bingxuan and Wu, Bochao and Lu, Chengda and Zhao, Chenggang and Deng, Chengqi and Zhang, Chenyu and Ruan, Chong and others},
  journal={arXiv:2412.19437},
  year={2024}
}

@article{jaech2024openai,
  title={Openai o1 system card},
  author={Jaech, Aaron and Kalai, Adam and Lerer, Adam and Richardson, Adam and El-Kishky, Ahmed and Low, Aiden and Helyar, Alec and Madry, Aleksander and Beutel, Alex and Carney, Alex and others},
  journal={arXiv:2412.16720},
  year={2024}
}

@article{schulman2017proximal,
  title={Proximal policy optimization algorithms},
  author={Schulman, John and Wolski, Filip and Dhariwal, Prafulla and Radford, Alec and Klimov, Oleg},
  journal={arXiv:1707.06347},
  year={2017}
}

@article{rafailov2023direct,
  title={Direct preference optimization: Your language model is secretly a reward model},
  author={Rafailov, Rafael and Sharma, Archit and Mitchell, Eric and Manning, Christopher D and Ermon, Stefano and Finn, Chelsea},
  journal={Advances in neural information processing systems},
  volume={36},
  pages={53728--53741},
  year={2023}
}

@article{zhang2025r1,
  title={R1-vl: Learning to reason with multimodal large language models via step-wise group relative policy optimization},
  author={Zhang, Jingyi and Huang, Jiaxing and Yao, Huanjin and Liu, Shunyu and Zhang, Xikun and Lu, Shijian and Tao, Dacheng},
  journal={arXiv:2503.12937},
  year={2025}
}

@article{zhang2025critique,
  title={Critique-grpo: Advancing llm reasoning with natural language and numerical feedback},
  author={Zhang, Xiaoying and Sun, Hao and Zhang, Yipeng and Feng, Kaituo and Lu, Chaochao and Yang, Chao and Meng, Helen},
  journal={arXiv:2506.03106},
  year={2025}
}

@article{ramesh2024group,
  title={Group robust preference optimization in reward-free rlhf},
  author={Ramesh, Shyam Sundhar and Hu, Yifan and Chaimalas, Iason and Mehta, Viraj and Sessa, Pier Giuseppe and Bou Ammar, Haitham and Bogunovic, Ilija},
  journal={Advances in Neural Information Processing Systems},
  volume={37},
  pages={37100--37137},
  year={2024}
}

@article{gao2024rebel,
  title={Rebel: Reinforcement learning via regressing relative rewards},
  author={Gao, Zhaolin and Chang, Jonathan and Zhan, Wenhao and Oertell, Owen and Swamy, Gokul and Brantley, Kiant{\'e} and Joachims, Thorsten and Bagnell, Drew and Lee, Jason D and Sun, Wen},
  journal={Advances in Neural Information Processing Systems},
  volume={37},
  pages={52354--52400},
  year={2024}
}

@article{shao2024deepseekmath,
	title   = {Deepseekmath: Pushing the limits of mathematical reasoning in open language models},
	author  = {Shao, Zhihong and Wang, Peiyi and Zhu, Qihao and Xu, Runxin and Song, Junxiao and Bi, Xiao and Zhang, Haowei and Zhang, Mingchuan and Li, YK and Wu, Y and others},
	journal = {arXiv:2402.03300},
	year    = {2024}
}

@InProceedings{Liu_2025_ICCV,
    author    = {Liu, Ziyu and Sun, Zeyi and Zang, Yuhang and Dong, Xiaoyi and Cao, Yuhang and Duan, Haodong and Lin, Dahua and Wang, Jiaqi},
    title     = {Visual-RFT: Visual Reinforcement Fine-Tuning},
    booktitle = {Proceedings of the IEEE/CVF International Conference on Computer Vision (ICCV)},
    month     = {October},
    year      = {2025},
    pages     = {2034-2044}
}

@inproceedings{yang2010bag,
  title={Bag-of-visual-words and spatial extensions for land-use classification},
  author={Yang, Yi and Newsam, Shawn},
  booktitle={Proceedings of the 18th SIGSPATIAL international conference on advances in geographic information systems},
  year={2010}
}

@Article{Dai2011WHURS19,
title={Satellite Image Classification via Two-Layer Sparse Coding With Biased Image Representation},
author={Dengxin Dai and Wen Yang},
journal={IEEE Transactions on Geoscience and Remote Sensing},
year={2011},
volume={8},
number={1},
pages={173-176}
}

@article{cheng2017remote,
  title={Remote sensing image scene classification: Benchmark and state of the art},
  author={Cheng, Gong and Han, Junwei and Lu, Xiaoqiang},
  journal={Proceedings of the IEEE},
  volume={105},
  number={10},
  pages={1865--1883},
  year={2017},
  publisher={IEEE}
}

@article{zhou2018patternnet,
  title={PatternNet: A benchmark dataset for performance evaluation of remote sensing image retrieval},
  author={Zhou, Weixun and Newsam, Shawn and Li, Congmin and Shao, Zhenfeng},
  journal={ISPRS journal of photogrammetry and remote sensing},
  volume={145},
  pages={197--209},
  year={2018},
  publisher={Elsevier}
}

@article{xia2017aid,
  title={AID: A benchmark data set for performance evaluation of aerial scene classification},
  author={Xia, Gui-Song and Hu, Jingwen and Hu, Fan and Shi, Baoguang and Bai, Xiang and Zhong, Yanfei and Zhang, Liangpei and Lu, Xiaoqiang},
  journal={IEEE Transactions on Geoscience and Remote Sensing},
  volume={55},
  pages={3965--3981},
  year={2017},
  publisher={IEEE}
}

@article{lu2017exploring,
  title={Exploring models and data for remote sensing image caption generation},
  author={Lu, Xiaoqiang and Wang, Binqiang and Zheng, Xiangtao and Li, Xuelong},
  journal={IEEE Transactions on Geoscience and Remote Sensing},
  volume={56},
  number={4},
  pages={2183--2195},
  year={2017},
  publisher={IEEE}
}

@article{medmnistv2,
    title={MedMNIST v2-A large-scale lightweight benchmark for 2D and 3D biomedical image classification},
    author={Yang, Jiancheng and Shi, Rui and Wei, Donglai and Liu, Zequan and Zhao, Lin and Ke, Bilian and Pfister, Hanspeter and Ni, Bingbing},
    journal={Scientific Data},
    volume={10},
    number={1},
    pages={41},
    year={2023},
    publisher={Nature Publishing Group UK London}
}

@article{li2020object,
  title={Object detection in optical remote sensing images: A survey and a new benchmark},
  author={Li, Ke and Wan, Gang and Cheng, Gong and Meng, Liqiu and Han, Junwei},
  journal={ISPRS journal of photogrammetry and remote sensing},
  volume={159},
  pages={296--307},
  year={2020},
  publisher={Elsevier}
}

@article{cao2025generalized,
  title={Generalized domain prompt learning for accessible scientific vision-language models},
  author={Cao, Qinglong and Chen, Yuntian and Lu, Lu and Sun, Hao and Zeng, Zhengzhong and Yang, Xiaokang and Zhang, Dongxiao},
  journal={Nexus},
  volume={2},
  number={2},
  year={2025},
  publisher={Elsevier}
}
\bibliographystyle{icml2026}

%%%%%%%%%%%%%%%%%%%%%%%%%%%%%%%%%%%%%%%%%%%%%%%%%%%%%%%%%%%%%%%%%%%%%%%%%%%%%%%
%%%%%%%%%%%%%%%%%%%%%%%%%%%%%%%%%%%%%%%%%%%%%%%%%%%%%%%%%%%%%%%%%%%%%%%%%%%%%%%
% APPENDIX
%%%%%%%%%%%%%%%%%%%%%%%%%%%%%%%%%%%%%%%%%%%%%%%%%%%%%%%%%%%%%%%%%%%%%%%%%%%%%%%
%%%%%%%%%%%%%%%%%%%%%%%%%%%%%%%%%%%%%%%%%%%%%%%%%%%%%%%%%%%%%%%%%%%%%%%%%%%%%%%
% \newpage
% \appendix
% \onecolumn

% You can have as much text here as you want. The main body must be at most $8$
% pages long. For the final version, one more page can be added. If you want, you
% can use an appendix like this one.

% The $\mathtt{\backslash onecolumn}$ command above can be kept in place if you
% prefer a one-column appendix, or can be removed if you prefer a two-column
% appendix.  Apart from this possible change, the style (font size, spacing,
% margins, page numbering, etc.) should be kept the same as the main body.
%%%%%%%%%%%%%%%%%%%%%%%%%%%%%%%%%%%%%%%%%%%%%%%%%%%%%%%%%%%%%%%%%%%%%%%%%%%%%%%
%%%%%%%%%%%%%%%%%%%%%%%%%%%%%%%%%%%%%%%%%%%%%%%%%%%%%%%%%%%%%%%%%%%%%%%%%%%%%%%

\end{document}